%% file: main.tex
\pdfoutput=1

\documentclass[11pt]{article}

\usepackage[]{EMNLP2022}

\usepackage{times}
\usepackage{latexsym}

\usepackage[T1]{fontenc}

\usepackage[utf8]{inputenc}
\usepackage{booktabs}
\usepackage{microtype}

\newcommand{\avivs}[1]{{\color[rgb]{0.7,0.3,0}{}}}

\usepackage{graphicx}
\usepackage{amsmath}

\usepackage{textcomp}

\usepackage{hyperref}

\usepackage{subcaption}

\usepackage{adjustbox}

\usepackage{enumitem}
\usepackage{tabularx}

\usepackage{inconsolata}

\title{Controlled Text Reduction}

\author{Aviv Slobodkin, Paul Roit, Eran Hirsch, Ori Ernst, Ido Dagan \\        
        Bar-Ilan University\\
        {\tt \{lovodkin93,plroit,hirsch.eran,oriern\}@gmail.com} \\
        \texttt{dagan@cs.biu.ac.il}}

\begin{document}
\maketitle

\input{sections/01_abstract}

\input{sections/02_introduction}

\input{sections/03_Background}

\input{sections/04_task_definition}

\input{sections/05_crowdsourcing}

\input{sections/06_train_dataset}

\input{sections/07_baseline_models}
\input{sections/08_Evaluation_and_Analysis}

\input{sections/09_conclusion}

\input{sections/10_limitations}

\input{sections/11_acknowledgements}


\bibliography{anthology,custom}
\bibliographystyle{acl_natbib}
\include{appendix}

\end{document}

%% file: sections/01_abstract.tex
\begin{abstract}
Producing a reduced version of a source text, as in generic or focused summarization, inherently involves two distinct subtasks: deciding on targeted content and generating a coherent text conveying it. While some popular approaches address summarization as a single end-to-end task, prominent works support decomposed modeling for individual subtasks. 
Further, semi-automated text reduction is also very appealing, where users may identify targeted content while models would generate a corresponding coherent summary.

In this paper, we focus on the second subtask, of generating coherent text given pre-selected content. Concretely, we formalize \textit{Controlled Text Reduction} as a standalone task, whose input is a source text with marked spans of targeted content ("highlighting").
A model then needs to generate a coherent text that includes all and only the target information.
We advocate the potential of such models, both for modular fully-automatic summarization, as well as for semi-automated human-in-the-loop use cases.
Facilitating proper research, we crowdsource high-quality dev and test datasets for the task. Further, we automatically generate a larger "silver" training dataset from available summarization benchmarks, leveraging a pretrained summary-source alignment model.
Finally, employing these datasets, we present a supervised baseline model, showing promising results and insightful analyses.\footnote{Our data and code are released for open access:\\ \url{https://huggingface.co/datasets/biu-nlp/Controlled-Text-Reduction-dataset} \\ \url{https://github.com/lovodkin93/Controlled_Text_Reduction}}
\end{abstract}

%% file: sections/02_introduction.tex
\section{Introduction}

\begin{figure*}[h!]
\centering
    \includegraphics[width=1\linewidth]{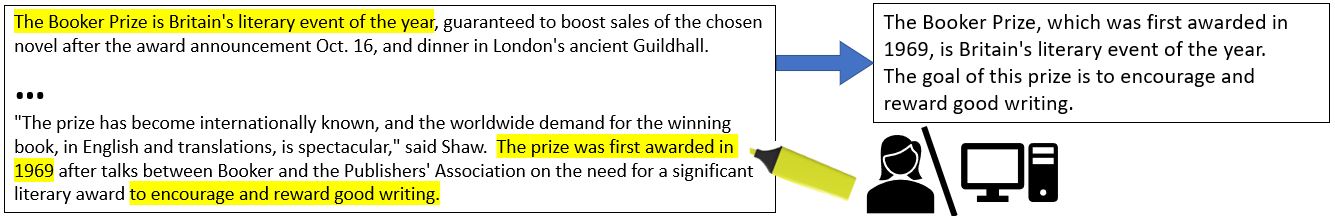}
    \caption{An example of an input, consisting of a source document and highlights (left), and the generated passage covering the highlighted content while preserving coherence (right). Such highlights in realistic use cases may be produced either by a human user or by a salience detection model.}
    \label{fig:example_task}
    \vspace{-0.2cm}
\end{figure*}

Abstractive text summarization takes one or more documents as input and aims at generating an accurate and coherent summary from it. It requires both locating salient information in the input and then generating a concise text covering it. 
While some modern state-of-the-art abstractive summarization models treat the task as a single end-to-end task, it has been common practice for summarization models to separate the salience detection phase from the text generation phase \citep{barzilay-mckeown-2005-sentence, oya-etal-2014-template, https://doi.org/10.48550/arxiv.1609.07034, 10.1007/978-3-319-64206-2_54}, with renewed popularity in recent years \citep{lebanoff-etal-2019-analyzing, lebanoff-etal-2020-cascade, lebanoff-etal-2020-learning, XIAO2022108483, https://doi.org/10.48550/arxiv.2112.08770, gehrmann-etal-2018-bottom, chen-bansal-2018-fast, cho-etal-2019-improving}. But, though those proposed techniques comprised distinguishable subtasks, evaluation was performed on the whole summarization pipeline, rather than optimizing each step separately.


In this paper, we focus on the text generation step, while addressing it as a standalone task at the sub-sentence level.
To that end, we introduce a new task which we denote \textit{Controlled Text Reduction}. The task takes as input a document with pre-chosen salient spans in it, which we will henceforth call \textit{highlights}. A model is then expected to reduce the document to a smaller coherent text which covers all and only the highlighted content, i.e., consolidating the highlighted spans into a fluent and coherent passage, as exemplified in \autoref{fig:example_task}. 
This task poses a challenge, as it requires generating fluent and grammatical text from non-consecutive spans while keeping it faithful to the source document. Hence, to balance the coherency and faithfulness constraints, models will be expected to use the context document to fill in implied details and to properly connect the different spans.

Focusing on this task can facilitate greater control over the generated text. It could lead to a modular summarization pipeline, where text-generation models can be trained once, and then used with different content selections to accommodate different needs. For example, we may envision a user (e.g., a student) pre-selecting the desirable textual content (either manually or via a designated model) while focusing on personal needs, possibly interactively \citep{hirsch-etal-2021-ifacetsum, shapira-etal-2021-extending}. Then, an available controlled text reduction module would transform the pre-selected fragments into a concise summary.
Also, separating the content selection and generation stages can lead to developing data-efficient systems, one to model salient content and another to generate the text. It could also lead to a more efficient characterization and research of each step separately without the need for probing, which is the prevailing approach in end-to-end models \cite{conneau-etal-2018-cram, tenney-etal-2019-bert, tenney2018what, slobodkin-etal-2021-mediators, pandit-hou-2021-probing}.

To promote research on the advocated text reduction task, we first develop a suitable controlled crowdsourcing methodology, following \citet{roit-etal-2020-controlled}, and apply it to produce high-quality dev and test datasets (\S\ref{sec:Crowd Sourcing}). Next, we automatically generate a larger training dataset,  by aligning propositional units of information \citep{ernst-etal-2021-summary}, extracted with OpenIE \citep{stanovsky-etal-2018-supervised}, between source documents and their summaries (\S\ref{sec:Train Dataset}). We use this data to train an abstractive supervised model, and evaluate its performance against our testset while comparing it to an extractive reference baseline, which simply concatenates the highlights. We also perform analyses where we manipulate the highlights and show that the addition of highlights to a supervised model is helpful in steering the model toward the pre-selected content, in addition to improving overall faithfulness and fluency (\S\ref{subsec:Results}).

Hence, the contribution of this paper is manifold:
\begin{enumerate}[topsep=0pt,itemsep=-1ex,partopsep=1ex,parsep=1ex]
    \item Proposing the \textit{"Controlled Text Reduction"} task as a standalone module in automated or semi-automated use cases.
    \item Defining an intuitive and easy-to-reproduce crowd-sourcing method for the task.
    \item Constructing the first data suite for the task, including crowd-sourced dev and test sets and an automatically-generated train set.
    \item Developing a supervised baseline model for future work.
\end{enumerate}

%% file: sections/03_Background.tex
\section{Background}

\begin{figure*}[h!]
\centering
    \includegraphics[width=16cm]{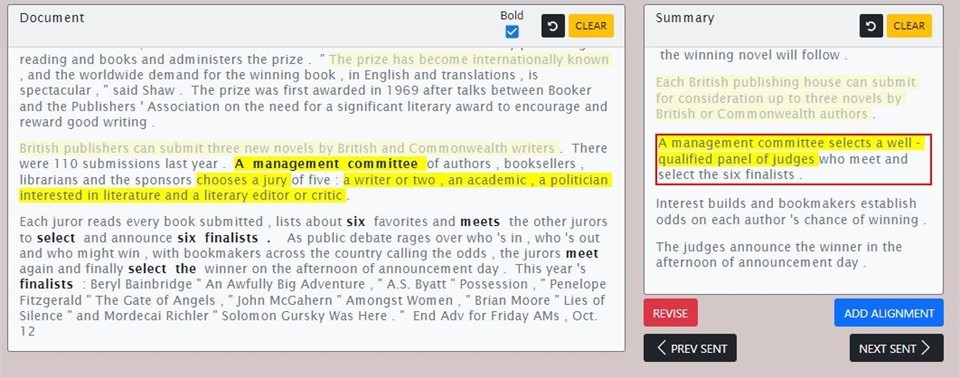}
    \caption{The Highlighting Annotation UI, presenting a document and its corresponding summary. Saved alignments have a faded yellow background, whereas currently selected alignments (which haven't been saved yet) have a normal yellow background. The current summary sentence is marked in a red box. Also, the bold feature is activated, meaning the document words which are related to those in the summary sentence are boldfaced (see \S\ref{subsec:Annotation Process}).}
    \label{fig:UI_example}
    \vspace{-0.2cm}
\end{figure*}

In this section, we briefly review related work and discuss the limitations of their framing.

As mentioned above, much of the related previous work focused primarily on end-to-end summarization \cite{carbonell1998use, haghighi2009exploring, nallapati2016classify, nallapati2016abstractive, paulus2017deep, gehrmann2018bottom}, with the vast majority of related datasets aimed at end-to-end summarization \cite{fabbri-etal-2019-multi, kim-etal-2019-abstractive, ghalandari2020large}, with only a source document as input. On the other hand, research on leveraging control through the injection of pre-chosen (rather than learned) signals in the seq-to-seq scenario focused mostly on semantic and syntactic signals, and also almost exclusively targeted Machine Translation models \citep{bugliarello-okazaki-2020-enhancing, akoury-etal-2019-syntactically, sundararaman2019syntaxinfused, choshen2021transition, slobodkin-etal-2022-semantics}. 

Attempts to leverage some control over the generation step in summarization received attention in recent years in the form of query-focused summarization \cite{baumel2018query, xu2020coarse, xu2021text, wei2017query} and keywords-focused summarization \cite{https://doi.org/10.48550/arxiv.1909.05858, https://doi.org/10.48550/arxiv.2012.04281}, with a few recently published corresponding datasets \cite{pasunuru2021data, kulkarni2020aquamuse, baumel2016topic}. A similar trend tried to leverage control through the addition of a planning step \citep{zhao-etal-2020-bridging, narayan-etal-2021-planning}.
Although these lines of research allowed for some control over salience, this control was limited and mostly focused on biasing the summary's topic, style, or structure.

The prevailing way to treat summarization in earlier works was to separate the salience detection phase from the text generation phase \citep{barzilay-mckeown-2005-sentence, oya-etal-2014-template, https://doi.org/10.48550/arxiv.1609.07034, 10.1007/978-3-319-64206-2_54}, yet the evaluation was performed on the whole pipeline. Some recent work focused on salience detection \citep{https://doi.org/10.48550/arxiv.2112.08770, ernst-etal-2021-summary, gehrmann-etal-2018-bottom, chen-bansal-2018-fast, cho-etal-2019-improving}, whereas the generation step has mostly been explored in a full-sentence-fusion setting \cite{geva-etal-2019-discofuse, lebanoff-etal-2019-analyzing, lebanoff-etal-2020-learning, XIAO2022108483}, rather than in a sub-sentence level. \citet{lebanoff-etal-2020-cascade} took it one step further, leveraging sentence fusion through a fine-grained content selection algorithm. But, though they did perform some analysis of this additional step by comparing different salience detection strategies, his evaluation focused on the full pipeline, similarly to his predecessors.


There has also been some work on extracting salient information in source documents in the form of highlights \cite{cho-etal-2020-better, arumae-etal-2019-towards}. Yet, though acknowledging the full potential of using highlights to mark salient information in the source document, it mainly focused on the process of obtaining these highlights, overlooking its actual usage in subsequent generation tasks, and in summarization in particular. Moreover, these lines of work focused solely on automatic highlight detection, lacking any crowdsourced annotation scheme. There has also been work that pre-identified salient parts as input to the generation phase \citep{chen-bansal-2018-fast, xu-etal-2020-self, liu-etal-2021-highlight, https://doi.org/10.48550/arxiv.2111.07935}
But, contrary to our work, the salience detection and generation tasks were addressed and evaluated jointly, without assessing the quality of each individual task.


All those research directions recognized the potential of separating the summarization task into subtasks and performing each subtask explicitly. However, they all evaluated the subtasks jointly, and in doing so overlooked the potential laying in the optimization and characterization of each task individually, and specifically the generation task given content-selection. In this work, we propose to isolate the generation task given pre-selected content, treating it as a stand-alone task, thus promoting focused evaluation and model designing.


%% file: sections/04_task_definition.tex
\section{Task Definition}

 \begin{figure*}[h!]
\centering
    \includegraphics[width=16cm]{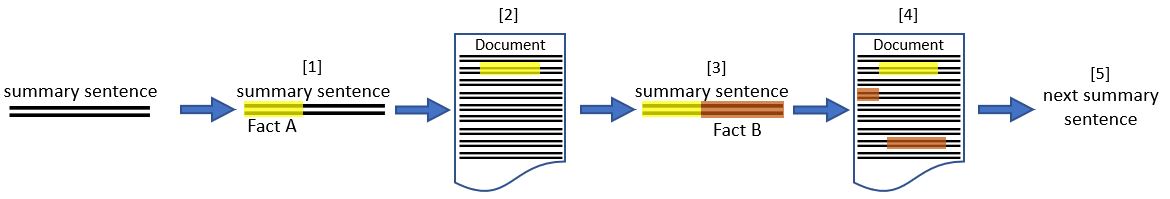}
    \caption{Illustration of Highlighting Annotation process for a summary sentence: [1] A \textit{summary} fact is located and highlighted; [2] The matching \textit{document} spans are highlighted, and the alignment is saved; [3] Another \textit{summary} fact is identified and highlighted; [4] The matching \textit{document} spans are highlighted, and the alignment is saved; [5] When the summary sentence is fully highlighted, we proceed to the next sentence, and so on. In this example, the summary consists of two facts, but steps 1 and 2 can be repeated as needed per sentence, until all its propositions (facts) are covered.}
    \label{fig:annotation_pipeline}
    \vspace{-0.2cm}
\end{figure*}

We define the \textit{controlled Text Reduction} task as follows. Given a document and a set of marked spans within that document, denoted as \textit{highlights}, produce a coherent output text encompassing only the information provided within those highlights (see \autoref{fig:example_task}).
The desirable output should adhere to two requirements beyond coherency: 
(1) Its content has to be derived from the highlights alone, keeping any additional document premises to the minimum required for coherency; (2) The output has to retain all of the details covered by the highlighted spans.

Such requirements give rise to many interesting challenges, such as recognizing the connecting thread between disparate spans and faithfully representing the information contained within them. 
We forgo a strict definition for a highlighted span and allow possibly marking sub-sentence elements: an entity or a clause, even discontinuous descriptions of these (e.g., the last two highlights in \autoref{fig:example_task}). 
Hence, the input highlights may be disconnected in both their surface realization (i.e. grammatically unsuitable), and semantic fluency.

\autoref{fig:example_task} features an input-output example. The output covers exclusively and completely the highlighted information while using the source document's context to connect the disparate spans.

%% file: sections/05_crowdsourcing.tex
\section{Gold Dataset for Evaluation}\label{sec:Crowd Sourcing}

We leverage different summarization datasets to annotate a high-quality dataset for the evaluation of controlled-reduction systems.
In summarization, every summary arises from a set of salient document spans.
Exploiting this in our annotation process, we wish to "reverse-engineer" each summary and locate the spans in the document that led to its construction. 
This significantly reduces the annotation complexity and load, instead of compiling a new text given a set of highlighted spans, an annotator has to highlight document spans given the output text (i.e. the summary).

\input{tables/dataset_statistics}

To create our development and test partitions we sample 121 and 108 unique documents from DUC 2001 and 2002 Single-Document-Summarization (SDS) datasets\footnote{\url{https://duc.nist.gov/}} respectively.
Each document is accompanied by up to 4 different reference summaries (with an average of 2.14 summaries per document), resulting in a total of 488 unique document-summary pairs (see \autoref{tab:dataset-statistics} for full statistics and \S\ref{sec:preprocessing} for preprocessing details).

We build an intuitive and convenient annotation tool for extracting highlights from document-summary pairs \footnote{\url{https://github.com/lovodkin93/highlights-extract-app}}, designed to be embedded into crowdsourcing platforms (see \S\ref{subsec:Annotation Process} and \autoref{fig:UI_example}).
Given the complexity of our task, we follow \citet{roit-etal-2020-controlled}'s \textit{controlled crowdsourcing} setup, including principled steps of annotator recruitment and training, leading to a trusted and qualified annotators group, employed for the annotation process.




\subsection{Annotation Process}\label{subsec:Annotation Process}

To annotate document spans, whose content corresponds to the summary content, we build a web-based user interface that is published on Amazon Mechanical Turk\footnote{\url{www.mturk.com}} and used by crowd-workers (see \autoref{fig:UI_example}). 
An annotator is presented with a document and its reference summary side-by-side and is instructed to highlight all of the phrases in the document whose content corresponds to the summary (see yellow background in \autoref{fig:UI_example}). 
To facilitate accurate and systematic processing of each instance, workers are asked to align spans from the summary that comprise a single fact to minimal spans in the document which cover them.
Thus, annotators create a series of alignments that cover every piece of information in the summary (see \autoref{fig:annotation_pipeline} for illustration of the annotation flow).

We observed that processing summary text one fact at a time substantially focuses the annotators' attention and expedites the search for relevant spans in the document.
This is exemplified when a single sentence in the summary is comprised of details that are mentioned in different locations spread out across the source document (e.g., the first summary sentence in \autoref{fig:example_task}). 
Further, to streamline the process, we segment the document into paragraphs and bolden content words in the document that share the same lemma with words in the current summary sentence (see document side in \autoref{fig:UI_example} and also \S\ref{sec:preprocessing} for details).
This method helps the human annotator to skim quickly through the document and is relatively bias-free. 
It is our assumption that a trained worker will not predominantly use same-lemma words for highlighting, as it is discouraged in our guidelines (see \S\ref{paragraph:Document-related Guidelines}).

After carefully assembling our trained worker pool, (see later \S\ref{subsec:Annotators Training}), each document-summary instance is annotated by a single worker. 
To supervise the resulting quality, we randomly sample submissions, supplying additional feedback if needed.


%

\subsection{Guidelines}\label{subsec:guidelines}

We instruct our workers to process the text systematically and align facts from each summary sentence to the corresponding phrases in the document. 

\paragraph{Summary-related Guidelines}
We provide guidelines for the annotator to break up the summary sentence into the facts that it is comprised of. We target facts encoded in main or embedded clauses, appositions, copular phrases, conjunctions, and more. \S\ref{subsec:Summary-related Guidelines} covers the full summary-related guidelines provided to the annotator.



\paragraph{Document-related Guidelines}\label{paragraph:Document-related Guidelines}
Once a summary fact was identified and highlighted, the crowd-workers are instructed to find its corresponding spans in the document. We define those spans as the minimal set of phrases that fully describe the current highlighted fact in the summary and nothing else. 
We define minimal in the sense that removing a content word from the document span would necessarily render some detail as not covered. For example, omitting anything from the first summary sentence in \autoref{fig:example_task}, e.g., "in 1969", would result in an overlooked highlighted fact.
Notably, the annotators may highlight multiple document spans portraying the same fact (redundantly in the document). Finally, we elaborate on the guidelines to touch down on issues such as paraphrasing, inconsecutive highlights, and highlighting in context. A more comprehensive overview of the guidelines and examples appears in \S\ref{subsec:Document-related Guidelines}.

\subsection{Annotator Training}\label{subsec:Annotators Training}
We follow the Controlled Crowdsourcing Methodology \cite{roit-etal-2020-controlled} to detect a group of qualified annotators, using two open qualification rounds for an initial selection, and proceeding with closed qualification rounds (for selected annotators) for further training and refining. 
In each round, the annotator is instructed to read a short description of the task and annotate a trial instance.\footnote{For the open rounds, the instance is simplified with a single summary sentence to focus on.} 
The closed qualification rounds proceeded with a 20-minute video explaining the different features of our annotation tool (see \S\ref{subsec:Annotation Process}).
Each round is followed by a thorough review of the authors for further feedback. 
The qualification rounds are fully paid, take up to 30 minutes to complete, and consist of 3 summary-document pairs and reading relevant feedback.
Upon completion, we remained with 11 annotators who successfully completed the training session, out of 15 who began the training round.


\paragraph{Cost}
We price every annotation instance, that takes on average 10 minutes to complete, at 2$\$$. 
We also compensate the workers for the time spent watching the 20-minute video during training with a 4$\$$ bonus upon completion of the video. The total dataset cost amounted to approximately 1400$\$$.

\subsection{Dataset Quality}\label{subsec: Dataset Quality}
To assess the quality of the resulting dataset we calculate different agreement scores between crowd-workers and experts. 
Given the same summary-document pair annotated separately by two annotators, we calculate Intersection-over-Union (\textit{IOU}) of the tokens' indices\footnote{We consider only content words.} between the highlighted document spans that are aligned to the same summary sentence, similarly to \citet{ernst-etal-2021-summary}.
We collect the sentence-wise IOU scores across 3 summary-document pairs, annotated by 11 workers to calculate the Inter-Annotator-Agreement and find that our workers exhibit a high agreement of 82.09, suggesting that our annotation protocol is well-defined and stable.
Likewise, we calculate the agreement between the annotators to references made by two of the authors and find it to be also high (78.23), indicating a good quality of our annotated data.

From analyzing all disagreements ($IoU<90\%$), we find that the main factor for disagreement stems from two separate spans in the document entailing the same event, resulting in each of the annotators highlighting a different mention of it or in one of them highlighting both mentions. This does not harm the quality of our data, as both options are fitting for the task. Another prevalent reason for disagreement arises from one of the annotators highlighting extra phrases that overall add only insignificant details on top of the summary. For examples, see \S\ref{sec:IAA disagreement Examples}. 
Finally, an interesting characteristic of our dataset is that for $>40\%$ our annotated data, a summary sentence is aligned with non-consecutive phrases originating in different document sentences (see \autoref{tab:dataset-statistics}), representing the challenges faced by a text reduction model in a realistic setting.

%% file: tables/dataset_statistics.tex
\begin{table*}[h!]
\centering
\begin{adjustbox}{width=1\textwidth}
\begin{tabular}{|l|l|l|l|l|l|l|l|}
\hline
 & \#unique docs & \begin{tabular}[c]{@{}l@{}}\#summaries/doc \\ (average)\end{tabular} & \begin{tabular}[c]{@{}l@{}}\#summary-doc \\ pairs\end{tabular} & \begin{tabular}[c]{@{}l@{}}mean input/output \\ size (tkns)\end{tabular} & \begin{tabular}[c]{@{}l@{}}max input/output \\ size (tkns)\end{tabular} & \begin{tabular}[c]{@{}l@{}}mean input/output \\ size (sentences)\end{tabular} & \begin{tabular}[c]{@{}l@{}}summary sentences aligning \\ to multiple doc sentences\end{tabular} \\ \hline
Train (DUC) & 893 & 2.14 & 1911 & 849.13/115.34 & 8311/153 & 35.73/4.60 & 41.87 \% \\ \hline
Dev (DUC) & 57 & 2.26 & 129 & 790.95/121.05 & 3079/164 & 27.68/4.44 & 40.62 \% \\ \hline
Test (DUC) & 172 & 2.09 & 359 & 876.35/120.59 & 3384/161 & 30.84/4.34 & 40.71 \% \\ \hline
Overall (DUC) & 1122 & 2.14 & 2399 & 850.40/116.44 & 8311/164 & 34.58/4.56 & 41.63 \% \\ \hline
Train (CNN-DM) & 285073 & 1 & 285073 & 810.77/56.91 & 2934/2100 & 40.07/2.72 & 71.29 \% \\ \hline
\end{tabular}
\end{adjustbox}
\caption{Statistics of our dataset, including the number of unique documents, the average number of summaries per document, the number of summary-document pairs (a unique document creates a pair with each of its summaries), the mean input/output size (in tokens and in sentences), the maximum input/output size (in tokens) and the percentage of sentences whose alignments span across more than one document sentence.}
\label{tab:dataset-statistics}
\end{table*}

%% file: sections/06_train_dataset.tex
\section{Train Dataset}\label{sec:Train Dataset}
To acquire a larger dataset for training supervised models, we opt for an automatic approach to extract highlights. 
For that, we employ the superPAL model \cite{ernst-etal-2021-summary}, a proposition-based summary-source alignment model trained on a sentence alignment dataset \cite{Copeck2005LeveragingP, copeck2006leveraging, copeck2007catch, copeck2008update} based on the Pyramid evaluation method \cite{nenkova2004evaluating}. The model extracts propositions from the document and the summary, and then uses a RoBERTa encoder fine-tuned on MNLI and augmented with a binary classification layer to determine which propositions are aligned. 

We run the pre-trained superPAL model on the SDS DUC 2001 and 2002 document-summary pairs that were not already manually annotated (see \S\ref{sec:Crowd Sourcing}), consisting of 1911 such pairs (see \autoref{tab:dataset-statistics}), and the pairs of the CNN-DM train split \citep{https://doi.org/10.48550/arxiv.1602.06023}, consisting of 285073 such pairs (see \autoref{tab:dataset-statistics}).
For each pair, we collect only document highlights with an alignment probability of 0.5 or more, similarly to \citet{ernst-etal-2021-summary}.
This way, we perform automatically the task that was manually performed in \S\ref{sec:Crowd Sourcing}.
\input{tables/superpal_vs_mturk_macro}

\subsection{Evaluation of Automatic Annotation}\label{subsec:Train Data Evaluation}
Next, we wish to assess the quality of the automatically-generated data, and especially its correlation to the manually annotated dataset. For that, we first use SuperPal to extract potential highlights in the document-summary pairs annotated by our annotators (see \S\ref{sec:Crowd Sourcing}). Next, for every data point, we compare all its automatically-extracted highlights with their crowd-sourced counterparts.

\autoref{tab:SuperPal_vs_mturk_macro} presents the tokenwise\footnote{We consider only content words.}
 macro-averaged precision, recall, and F1 values, with the crowd-sourced highlights as the gold data (the micro-averaged values show similar trends - see \S\ref{sec:Train Data Micro-Averaged Evaluation}).
These results suggest that our automatically-generated highlights cover a substantial portion of the highlights, with reasonable precision, making them useful for large-scale training. However, these figures also stress the necessity of our manual annotation for the dev and test sets.





%% file: tables/superpal_vs_mturk_macro.tex
\begin{table}[]
\centering
\begin{tabular}{ccc}
\hline
P     & R     & F1    \\ \hline
66.17 & 68.35 & 65.24 \\ \hline
\end{tabular}
\caption{Token-wise macro-averaged precision, recall, and F1 scores when comparing the manually and automatically annotated document-summary pairs (dev$\&$test).}
\vspace{-0.4cm}
\label{tab:SuperPal_vs_mturk_macro}
\end{table}

%% file: sections/07_baseline_models.tex
\section{Baseline Models}\label{sec:Baseline Models}

We experiment with two methods for the controlled text reduction task: 
a supervised model, whose input is the full document, supplemented with indications of the highlighted spans (\S\ref{subsec:Highlighting in Context}) and another supervised model that receives as input only a concatenation of the highlights, without the surrounding context (\S\ref{subsec:Highlighting Correction}). Both models are trained on our automatically-generated train dataset (\S\ref{sec:Train Dataset}). 


\subsection{Highlights in Context}\label{subsec:Highlighting in Context}
Considering the length requirements of our data (see \autoref{tab:dataset-statistics}), we opt for a model designated for long inputs. 
We employ the  Longformer Encoder-Decoder base model \citep[LED\textsubscript{base};][]{https://doi.org/10.48550/arxiv.2004.05150}, with the standard configurations.\footnote{For details, see this \href{https://colab.research.google.com/drive/12LjJazBl7Gam0XBPy_y0CTOJZeZ34c2v?usp=sharing\%23scrollTo\%3DkQOaX6eRJXkM\%23scrollTo=kPnNi_tWaklV}{colab notebook}.}
The Longformer is an adaption of BART \cite{lewis-etal-2020-bart} for longer inputs, replacing BART's encoder with a combination of a local and a (optional) global attention mechanism. The local attention, which comes in the form of a sliding window, is mostly used to build contextual representations, by enabling each token to attend to its neighbors. Alternatively, a global attention, which is given to a few pre-selected input tokens, enables those tokens to attend to all the tokens in the input (and not only its neighbors), and also allows all input tokens to attend to the global ones. 
LED has demonstrated state-of-the-art results when evaluated on the arXiv long document summarization dataset \cite{cohan-etal-2018-discourse}, making it a suitable choice for our experiments. We denote this model LED\textsubscript{H}.

\subsection{Only Highlights}\label{subsec:Highlighting Correction}
To demonstrate the necessity of the document context, we also train a variant of the LED model where the input consists of a concatenation of the supplied document spans, without the surrounding context.\footnote{Given this input is short, we also experimented with Pegasus, which showed comparable results on the dev set.} We denote it LED\textsubscript{only-H}. We use the same configurations as in \S\ref{subsec:Highlighting in Context} while omitting the global attention (given it is not needed in this setting). 

%% file: sections/08_Evaluation_and_Analysis.tex
\section{Experimental Setup}\label{sec:Evaluation and Analysis}
\paragraph{Baseline Models}\label{paragraph:Baseline LED Model}


We use our training dataset (\S\ref{sec:Train Dataset}) to finetune our two LED variants (\S\ref{sec:Baseline Models}). We employ the CNN-DM dataset together with our DUC trainset for initial fine-tuning, which is then followed by further finetuning on the DUC trainset alone. We avoid using the CNN-DM dataset in the latter finetuning phase since its quality is notably lower compared to the DUC dataset. Specifically, CNN-DM was generated automatically, in comparison to the expert-written summaries in DUC, and it consists of standalone bullet points, lacking the desired discourse properties and flow of natural text.
To avoid overfitting on the CNN-DM dataset, which is much larger than DUC, we experimented with using only fractions of the CNN-DM data. Optimal performance was achieved when using the full CNN-DM data for the initial finetuning of the LED\textsubscript{H} model (\S\ref{subsec:Highlighting in Context}), while for the LED\textsubscript{only-H}  model it was best to finetune only on the DUC data, avoiding the CNN-DM data altogether.

In the LED\textsubscript{only-H}, we preprocess our input, extracting the highlights and then using a dot (followed by a space) to separate spans originating in different sentences, and a white space otherwise.
To model the highlights in the LED\textsubscript{H} setting, we follow \citet{https://doi.org/10.48550/arxiv.2111.07935} and add to the vocabulary two special tokens, \textit{<highlight\_start>} and \textit{<highlight\_end>}, which are inserted as vectors into the source documents' embedding layer at the beginning and end of each highlighted span. 
Also, we combine LED's local attention with its global attention mechanism. As the global attention adds bias to the designated tokens, we mark all \textit{<highlight\_start>} and \textit{<highlight\_end>} tokens as global tokens. Our motivation stems from the assumption that allowing all the highlight tokens to attend to one another (through the symmetry of the global attention) will encourage the model to fuse the information they are attached to, under the assumption that the highlighted spans are related. Though LED supports inputs with up to 16384 tokens, for our purposes we limit it to 4096 tokens (see \autoref{tab:dataset-statistics}). 

We also examined other techniques to represent the highlights \citep{chen-bansal-2018-fast, xu-etal-2020-self, liu-etal-2021-highlight}, but as they introduced dependencies between their salience detection and generation components, we found them less fitting in our setting.

As a reference point, we compare the abstractive models to an extractive text generated by simply concatenating the highlights, as described previously (i.e., the input to LED\textsubscript{only-H}).
This version serves to demonstrate the necessity of our new abstractive task formulation, since without a system that bridges disparate texts, the concatenated spans are often unintelligible.

\paragraph{No Highlights} 
In addition to the two baseline models for our text reduction task, we also examine LED in a standard no-highlight summarization setting, where it is finetuned and evaluated on the original document without any highlights. 
In the absence of highlights, the global attention becomes unnecessary, hence this variant incorporates solely local attention. 
This no-highlight variant of LED, denoted LED\textsubscript{NH}, matches the classic summarization setting and provides insights into the ability of the model to pick up the highlighting signals.
When optimizing the amount of CNN-DM data to use in the initial finetuning phase, as described above for the baseline models, we found it optimal to use $5\%$ of the CNN-DM data.

\paragraph{Highlights-Summary Mix}
To investigate the extent of the highlights' impact, we create a variant of our highlighted test setting: 
For each document-summary pair, we assign highlighted spans that were extracted from another reference summary available for the same document.
We use all the document-summary pairs in our test set for this experiment.
We then evaluate the finetuned LED\textsubscript{H} (see \S\ref{subsec:Highlighting in Context}) on this setting, evaluating the generated summary for each highlighted document with the summary that may have different salient content than the highlights, denoting it LED\textsubscript{H-mix}. 
\paragraph{Manual Fluency Evaluation}\label{paragraph:Fluency Evaluation}
To test our assumption that simply concatenating the highlights, or excluding the document context, results in less coherent text, we ask crowd-workers to rate the fluency of the generated texts for the suggested baseline models. Our group of crowd-workers consists of reliable workers that have shown a good grasp of different semantic tasks including summarization in past experiments. To evaluate, we randomly choose 100 documents from our test set, each is assigned a single set of highlights corresponding to some summary. 
We design a simple Amazon Mechanical Turk interface, where we present all three generated summaries of the same input (see \S\ref{sec:Fluency Human Evaluation API}). 
Inspired by \citet{fabbri-etal-2021-summeval}'s judgment guidelines to crowd-workers, we use a 5-point Likert scale to evaluate the consistency and fluency of the generated summaries and add criteria explaining each score, to reduce ambiguity and enforce more consistent rating (see \S\ref{sec:Fluency Human Evaluation API}).

\input{tables/fluency_evaluation}

\section{Analysis and Results}\label{subsec:Results}
First, we present the fluency results to validate the necessity of our task setting. As expected, it arises from \autoref{tab:fluency_human_evaluation} that the Concat. approach generates highly incoherent summaries, as opposed to the supervised model. This shows that just copying from the highlights directly leads to incoherent text. We also see that removing the context from the input is also detrimental to the model's ability to generate a coherent text (LED\textsubscript{H} vs. LED\textsubscript{only-H}), demonstrating the importance of context (see \S\ref{sec:Generation Examples} for example generated texts). 
To obtain further insight into context importance, we manually inspect the crowd-sourced datasets and find that for $~74\%$ of the document-summary pairs, context is indeed required to properly connect the disparate highlighted spans.

Next, we proceed to evaluate content preservation using ROUGE \citep{Lin2003AutomaticEO}, a lexical overlap metric (see \autoref{tab:evaluation-results-compared-to-highlights}).
To measure content preservation we apply the metric between the generated text and the highlighted content aimed to be preserved (technically, the highlights are concatenated to apply the ROUGE measure).\footnote{It is worth mentioning that we observed similar trends when comparing the generated texts to the original summaries - see \S\ref{sec:ROUGE results When Compared to the Gold Summaries} for further details. }
As may expected, it arises from \autoref{tab:evaluation-results-compared-to-highlights} that passing only the highlights through a supervised model results in the best ROUGE scores (see LED\textsubscript{only-H}), suggesting that, in the absence of additional content, the LED\textsubscript{only-H} model tends to preserve the original lexical content within its input highlights. Yet, as was seen in \autoref{tab:fluency_human_evaluation}, avoiding the context yields unacceptably incoherent text, making this model irrelevant to the task. 
Adding context to the input (LED\textsubscript{H}) downgrades the ROUGE score, which may be attributed to either desired or undesired behaviors of the LED\textsubscript{H} model.
In some cases, the generated text does preserve the highlighted content, but deviates from it lexically in order to generate fluent text, possibly incorporating certain lexical elements from the context while preserving meaning.
In other cases, however, the generated text does deviate from the highlighted content by erroneously adding to the output non-highlighted content from the surrounding context. Unfortunately, the ROUGE measure, being based solely on lexical matches, does not distinguish between these two cases. To that end, we add a manual faithfulness analysis in \S\ref{subsec:Performance Analysis} (\autoref{tab:Manual_Analysis_micro}), which evaluates content preservation more precisely, with respect to both precision (faithfulness) and recall (coverage).




Finally, we observe an approximately 8 points decrease in all ROUGE metrics when removing the highlights (LED\textsubscript{NH}), indicating that highlights do in fact play a major role in directing the model to focus on specific targeted content. We see a similar trend in LED\textsubscript{H-mix}, suggesting that each set of highlights steers the model toward the specific content it focuses on. 
This further confirms the highlights' role in the model's content-related decisions.

In conclusion, to evaluate future progress on the text reduction task, we firstly propose combining manual evaluation of fluency, requiring sufficient fluency to make models acceptable, along with automatic evaluation of content preservation via common measures for this purpose such as ROUGE. 
While we also inspected less standard automatic evaluation measures, for both fluency \citep{mutton-etal-2007-gleu} and semantic-oriented content matching \citep{honovich-etal-2021-q2, laban-etal-2022-summac}, we found them to be not sufficiently reliable for our setting.
That said, future progress in the quality of automatic evaluation of summary fluency and content matching would be highly applicable, and desired, for our text reduction task as well, particularly given the known deficiencies of the lexical-matching-based ROUGE measure. Further, reliable crowdsourcing methods for human evaluation of content matching may be considered as well \citep{shapira-etal-2019-crowdsourcing}, as we illustrate in our limited-scale analysis in the next subsection.

\subsection{Performance Analysis}\label{subsec:Performance Analysis}

\input{tables/evaluation_results_compared_to_highlights}

\input{tables/manual_analysis}

To further evaluate the highlights' effect, we manually assess LED\textsubscript{H} and LED\textsubscript{NH} on two levels: (1) faithfulness of the generated text and (2) coverage of the highlighted spans in the system summary.

To determine the amount of system summary spans that are entailed by the source, we compare each summary span to the source. We conducted two manual experiments, one with respect to the full document, and one with respect to the highlighted spans only. To that end, we randomly select 10 unique documents from our test set, with one of their set of highlights. 
Then, following the notion of Summary Content Unit (SCU) in the Pyramid method for summarization evaluation \citep{nenkova-passonneau-2004-evaluating}, we extract such units from both the summary and the source text using the Summary Evaluation Environment (SEE) described in that paper. Then, for each summary unit, we manually search for a matched document unit conveying the same information, to determine whether the summary unit is mentioned in the document (TP) or not (FP).
Lastly, we calculate the micro-precision, which represents the faithfulness of both models' outputs. \autoref{tab:Manual_Analysis_micro} shows an almost $5\%$ improvement in faithfulness to the source document when adding highlights. This implies that the highlights not only steer the model towards specific content but also help it keep focused on the source.
Interestingly, we find that one-third of the faithfulness errors (FP) stem from disparate highlights that were incorrectly combined, which is typical for summarization hallucinations.

We also evaluate the highlights' coverage by the summaries. For that, we calculate the number of False Negative (FN) summary facts, compared to the facts in the highlights, and compute the micro-recall value, representing the summaries' coverage of the highlights. 
\autoref{tab:Manual_Analysis_micro} shows a clear advantage to including highlights, with almost twice as big faithfulness (P) and coverage (R) of the highlighted facts. With that said, we note that the highlight-related faithfulness is still only a little over 50$\%$, indicating that the model included non-highlighted facts, which further exhibits the challenge to devise models that better focus only on the highlights.

%% file: tables/fluency_evaluation.tex
\begin{table}[]
\centering
\begin{tabular}{lll}
\hline
Concat.                  & LED\textsubscript{only-H} & LED\textsubscript{H} \\
\multicolumn{1}{c}{2.76} & \multicolumn{1}{c}{3.12}                       & \multicolumn{1}{c}{\textbf{4.58}}     \\ \hline
\end{tabular}
\caption{The (averaged) human ratings of fluency of the summaries generated by our two baseline models and the extractive reference model (Concat.).}
\label{tab:fluency_human_evaluation}
\vspace{-0.4cm}
\end{table}

%% file: tables/evaluation_results_compared_to_highlights.tex

\begin{table}[]
\begin{tabular}{|l|rrr|}
\hline
 & \multicolumn{1}{l}{R-1} & \multicolumn{1}{l}{R-2} & \multicolumn{1}{l|}{R-L} \\ \hline
LED\textsubscript{only-H} & \textbf{79.37} & \textbf{66.71} & \textbf{69.74} \\
LED\textsubscript{H} & 70.15 & 53.14 & 57.87 \\ \hline
LED\textsubscript{NH} & 49.98 & 28.89 & 36.55 \\
LED\textsubscript{H-mix} & 67.17 & 49.40 & 55.62 \\ \hline
\end{tabular}
\caption{ROUGE-1, -2 and -L content preservation results, comparing model output to the (concatenated) highlights in the input. We evaluate our baseline models (LED\textsubscript{H} and LED\textsubscript{only-H}), along with the alternative compared configurations (LED\textsubscript{NH} and LED\textsubscript{H-mix}).}
\label{tab:evaluation-results-compared-to-highlights}
\end{table}

%% file: tables/manual_analysis.tex
\begin{table}[]
\resizebox{\columnwidth}{!}{%
\begin{tabular}{|c|c|c|c|}
\hline
                               &                                        & Faithfulness (P) & Coverage (R)   \\ \hline
\multicolumn{1}{|l|}{Full Doc} & LED\textsubscript{NH} & 80.89            & N/A            \\ \hline
                               & LED\textsubscript{H}  & \textbf{85.11}   & N/A            \\ \hline
Highlights                     & LED\textsubscript{NH} & 29.94            & 27.33          \\ \hline
                               & LED\textsubscript{H}  & \textbf{52.48}   & \textbf{45.68} \\ \hline
\end{tabular}%
}
\caption{Fact-wise faithfulness (P) and coverage (R) scores for LED\textsubscript{NH} and LED\textsubscript{H}, once between generated summaries and the full source document and once between the generated summaries and the highlight.}
\vspace{-0.4cm}
\label{tab:Manual_Analysis_micro}
\end{table}

%% file: sections/09_conclusion.tex
\section{Conclusion}\label{sec:Conclusion}
In this paper, we promote the separation of the summarization task into the salience-detection and text-generation steps. 
We foresee applications where salient phrases will be highlighted by an avid reader, or selected by a model specialized in some domain, while a more general-purpose model would reformulate the disparate pieces into a coherent text.
Thus, we argue that \textit{Controlled Text Reduction}, the second step of summarization, is an interesting and useful research goal in its own right.
To bolster the task, we release a high-quality evaluation dataset and a heuristically-generated training data, evaluation protocol, and the first baseline model.
The latter clearly shows how the generated summary text benefits from the added salient span signals. 
Future works may expand this to include multi-document settings in order to accommodate the task to a broader range of applications, and also focus on designing better evaluation metrics for the task.



%% file: sections/10_limitations.tex
\section{Limitations}\label{sec:Limitations}
In this work, we construct the first-of-its-kind Controlled Text Reduction dataset, by aligning text spans in existing summaries to their correlated document spans. 
This poses a limitation on the highlights chosen, whereas in a more general setting users are free to highlight whatever they find interesting. On the contrary, in our setting, the highlights contain general salient information (that was extracted by the former human summarizer) rather than specific details. 

Also, our train dataset was derived automatically using the SuperPAL model. Hence, it is likely that some of the highlights in the training dataset are not perfectly aligned with the summary.

Finally, the dataset is based on a news corpus, which might limit its applicability to other applications that have different structures, such as medical or legal documents, or meeting summaries. 

%% file: sections/11_acknowledgements.tex
\section*{Acknowledgements}
This work was supported by Intel Labs, the Israel Science Foundation (grant no. 2827/21), and a grant from the Israel Ministry of Science and Technology.

%% file: appendix.tex
\appendix
\section{Preprocessing}\label{sec:preprocessing}
In preprocessing, we begin by removing meaningless characters from the input. Then, we use spaCy \cite{spacy2} to parse the input and the reference summaries to get their token segmentation, sentence separation, and lemmatization. Next, we construct a matrix $M_{ij}$ for each document-summary pair:

    \begin{equation}
    M_{ij} =  
    \begin{cases}
        1,& \text{$Similarity_{Lemma}(t^s_i, t^d_j)\geq0.86$}\\
        0,              & \text{otherwise}
    \end{cases}
    \end{equation}\label{eq:relation_matrix}

where $t^s_i$ and $t^d_j$ are \textbf{summary} token $i$ and \textbf{document} token $j$, respectively, and the $Similarity_{Lemma}(t^s_i, t^d_j)$ is computed using the SequenceMatcher\footnote{https://github.com/python/cpython/blob/main/Lib/difflib.py} module on $t^s_i$'s and $t^d_j$'s lemmas.

In addition, given that most of our dataset was not segmented into paragraphs, we devise a naive algorithm to divide the source documents of each data point in the dev and test datasets, in order to make them more presentable for our annotators and easier to read through. For that, we first apply the neuralcoref model\footnote{https://github.com/huggingface/neuralcoref} on the documents to get co-reference clusters, which we used together with the spaCy sentence segmentation to determine when paragraph-breaks should occur.




\section{Annotation Full Guidelines}\label{sec:Annotation Full Guidelines}
In this section, we provide the full annotation guidelines, presented to our workers.
\subsection{Summary-related Guidelines}\label{subsec:Summary-related Guidelines}
As mentioned in \ref{subsec:guidelines}, we provide guidelines for the annotator to segment the summary sentence into the facts that it is composed of. We target facts encoded in different grammatical structures, but to present them to the annotator in a simplified manner we show the following three variants: \\
$\bullet$ \textsc{side-by-side}: Two events are realized adjacently without sharing participants (e.g., \textit{"He worked"}, comprising of two independent events - \textit{"He worked while I slept"} and \textit{"I slept"}).\\
$\bullet$ \textsc{shared elements}: Two events that share some phrases (e.g., \textit{"He worked while smiling"}, which comprises of two events sharing a subject - \textit{"He worked"} and \textit{"He smiled"}). \\
$\bullet$ \textsc{no explicit verb}: An event is expressed without an explicit verb (e.g., \textit{"John Doe, my good friend, has arrived"}, whose first fact, \textit{"John Doe (is) my good friend"}, lacks an implicit verb).

\section{Document-related Guidelines}\label{subsec:Document-related Guidelines}
In this section, we present a more in-depth overview of the document-related guidelines presented to our annotators during their training.\footnote{The full guidelines presented during training upon publication.}: \\
- \underline{Paraphrasing}: We instruct our workers to not solely rely on phrases with shared words, as often the most suiting document phrase is a paraphrasing of its summary counterpart (for example, in \autoref{fig:UI_example}, "a well-qualified panel of judges" is a paraphrasing of its document counterpart). \\
- \underline{Consecutiveness}: We guide our workers to avoid highlighting unnecessary details, i.e., that did not appear in the summary span, and keep the highlights inconsecutive if necessary; (e.g., in \autoref{fig:UI_example}, the nature of the committee's members was excluded from the highlight, to adjust to the summary span, resulting in a non-consecutive highlight). \\
- \underline{Missing Details}: When the corresponding document phrase is missing some details, the annotators are instructed to highlight some other mention of the absent information. For example, in \autoref{fig:example_task}, the equivalent document span of the summary fact \textit{"The Booker Prize, which was first awarded in 1969"} is "The prize was first awarded in 1969". But, as the prize's "identity" is absent from this span, some mention of it should be highlighted as well (e.g., at the beginning of the document). \\
- \underline{hallucination}: For the rare instances where the reference summary has hallucinations, we instruct our workers to leave these details unhighlighted in the summary.\\
- \underline{Context}: We guide our workers to verify that the document highlights are used in the same context as the summary spans. For example, if in \autoref{fig:example_task} there was a mention of another prize that was awarded in 1969, highlighting it would be erroneous.


\section{IAA disagreement Examples}\label{sec:IAA disagreement Examples}
\begin{figure*}[t!]
    \centering
    \begin{subfigure}[h]{1.0\textwidth}
       \includegraphics[width=1\linewidth]{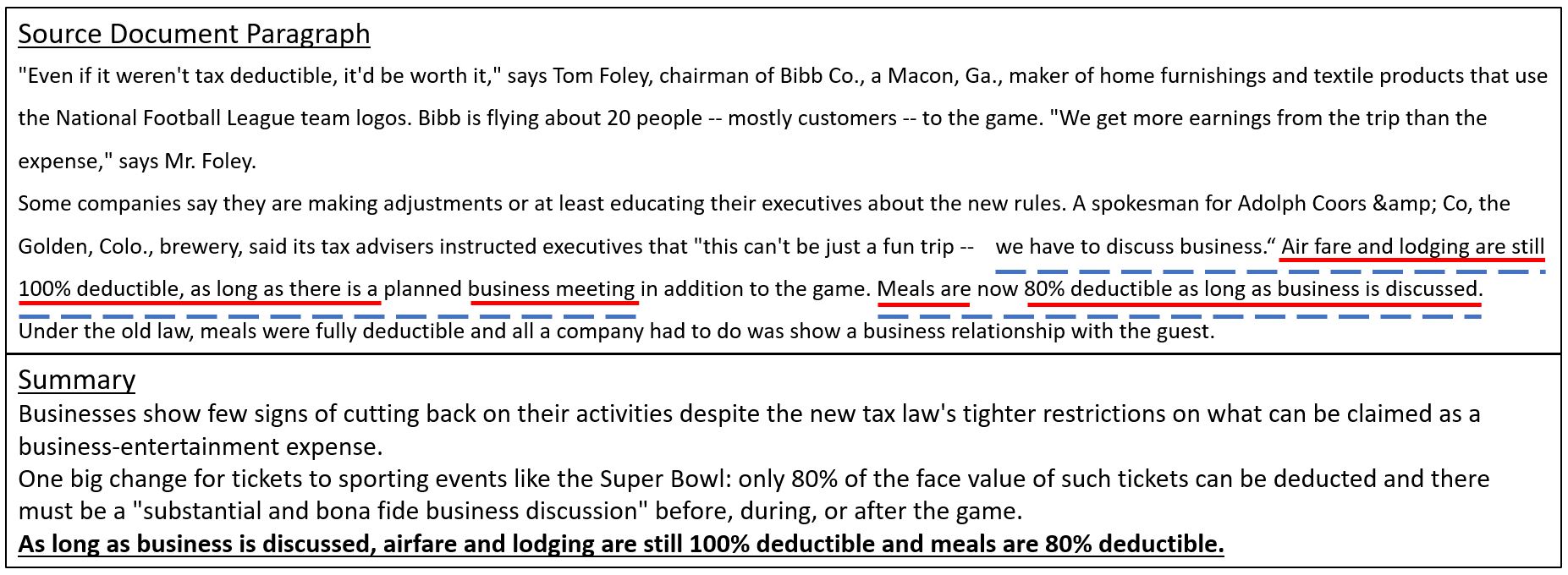}
       \caption{}
       \label{fig:IAA_disagreement_example0} 
    \end{subfigure}
    
    \begin{subfigure}[h]{1.0\textwidth}
       \includegraphics[width=1\linewidth]{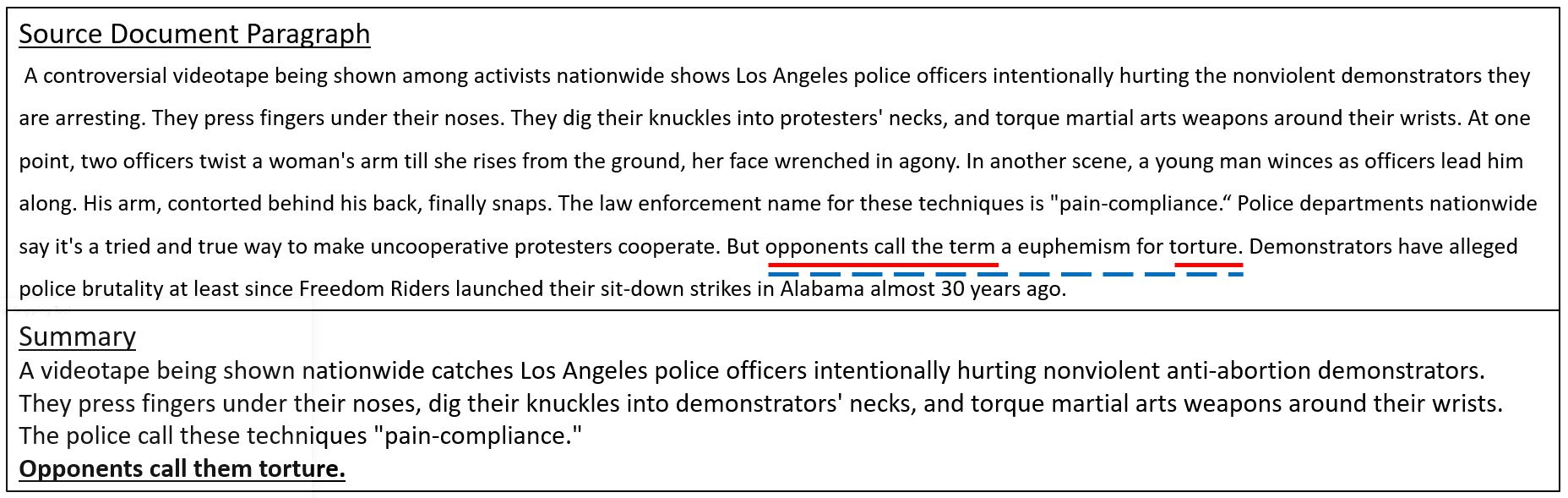}
       \caption{}
       \label{fig:IAA_disagreement_example1}
    \end{subfigure}
    
    \caption{Two examples of disagreement between annotators. For each example, the bottom part is the summary (with the summary sentence over which there was disagreement in bold and underlined) and the top part is a single paragraph from the source document with both the annotators' highlights (marked with a red solid line and a blue dashed line to indicate each highlight).}
    \label{fig:IAA_disagreement_examples}
\end{figure*}

\autoref{fig:IAA_disagreement_examples} exemplifies two disagreements between our annotators, which demonstrate the two main causes for disagreement (\S\ref{subsec: Dataset Quality}). In \autoref{fig:IAA_disagreement_example0}, we can see that one of the annotators highlighted an extra mention of the necessity to discuss business (dashed blue), which is allowed in our setup. In \autoref{fig:IAA_disagreement_example1}, we can see that one of the annotators included "a euphemism for" in the highlight (dashed blue), which has no effect on the overall meaning of the highlight.

\begin{figure*}
\centering
    \includegraphics[width=1\linewidth]{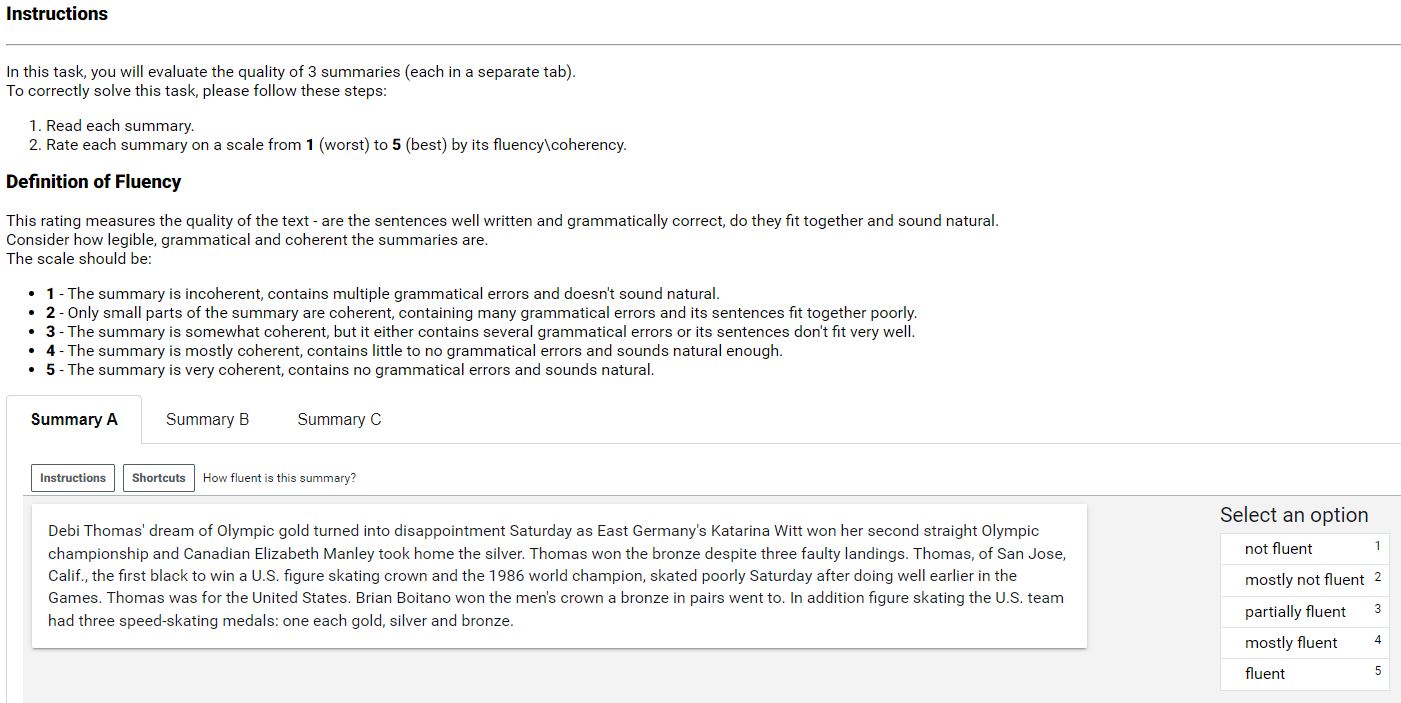}
    \caption{Example of the data collection API used by crowd-source workers.}
    \label{fig:fluency_evaluation_API_example}
\end{figure*}

\section{Train Data Micro-Averaged Evaluation}\label{sec:Train Data Micro-Averaged Evaluation}
\autoref{tab:SuperPal_vs_mturk_micro} shows the \textbf{micro}-averaged precision, recall, and F1 scores of the comparisons discussed in \autoref{subsec:Train Data Evaluation}.

\input{tables/superpal_vs_mturk_micro}

\section{Fluency Human Evaluation API}\label{sec:Fluency Human Evaluation API}
\autoref{fig:fluency_evaluation_API_example} present an example of our API designated for the human evaluation of the generated summaries' fluency and coherence. 

\section{Generation Examples}\label{sec:Generation Examples}
\begin{figure*}[h!]
\centering
    \includegraphics[width=16cm,height=24.5cm]{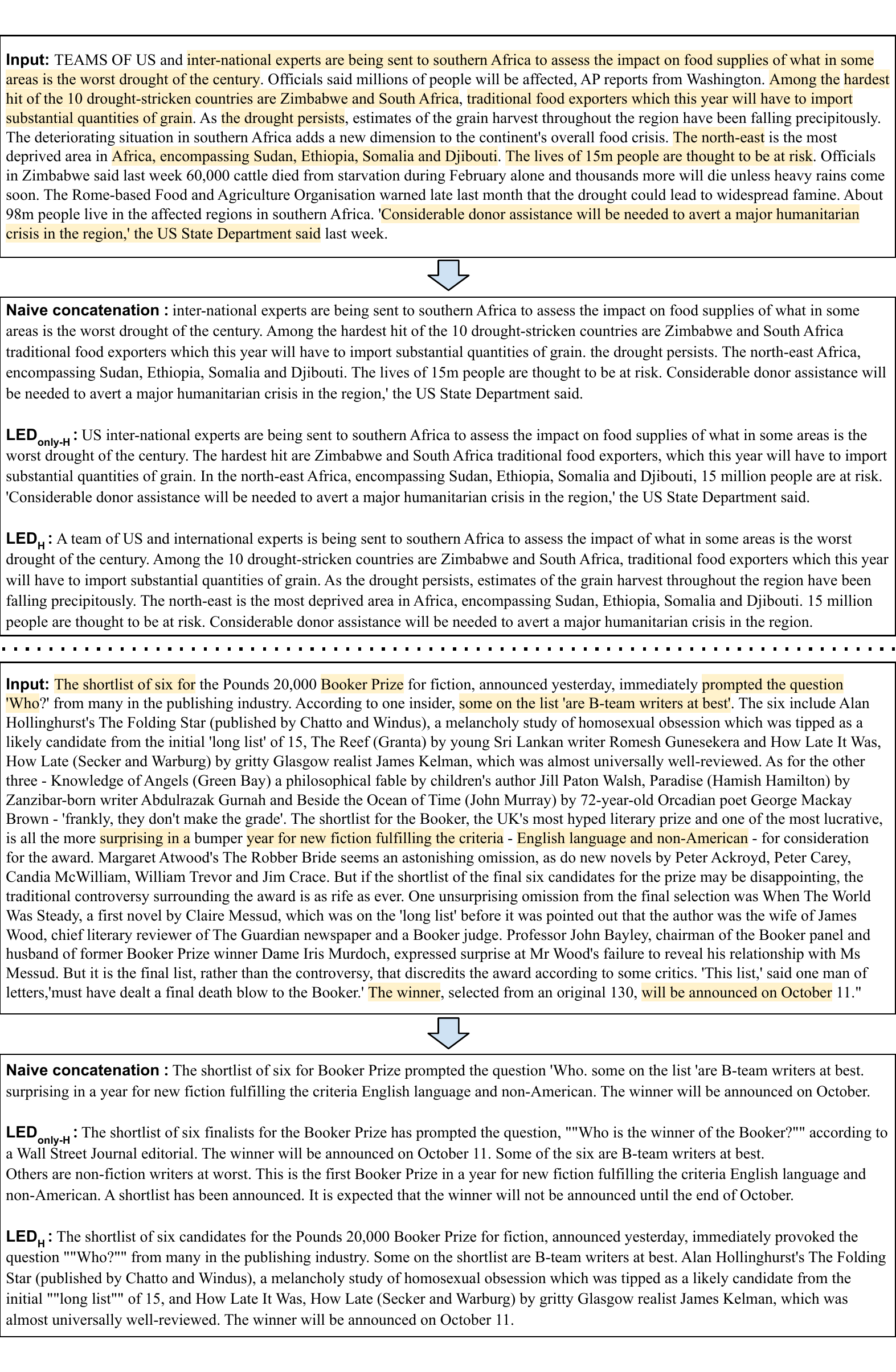}
    \caption{Example predictions from the various baseline models.}
    \label{fig:predictions}
\end{figure*}

Fig.~\ref{fig:predictions} shows two examples of a highlighted source document and the text generated by the Concat. approach (\textit{Naive concatentaion}) and our two baseline models.

\section{ROUGE results When Compared to the Gold Summaries}
\label{sec:ROUGE results When Compared to the Gold Summaries}
\autoref{tab:evaluation-results} features the ROUGE results of all our models (and also the Concat. extractive approach) when compared to the gold summaries.
\input{tables/evaluation_results}

%% file: tables/superpal_vs_mturk_micro.tex
\begin{table}[]
\centering
\begin{tabular}{ccc}
\hline
P     & R     & F1    \\ \hline
63.85 & 67.22 & 65.49 \\ \hline
\end{tabular}
\caption{Tokenwise micro-averaged precision, recall, and F1 scores when comparing the manually annotated document-summary pairs with the automatically-annotated pairs.}
\label{tab:SuperPal_vs_mturk_micro}
\end{table}

%% file: tables/evaluation_results.tex

\begin{table}[]
\begin{tabular}{|l|rrr|}
\hline
 & \multicolumn{1}{l}{R-1} & \multicolumn{1}{l}{R-2} & \multicolumn{1}{l|}{R-L} \\ \hline
Concat. & \multicolumn{1}{c}{\textbf{71.53}} & \multicolumn{1}{c}{\textbf{47.11}} & \multicolumn{1}{c|}{\textbf{52.53}} \\ \hline
LED\textsubscript{only-H} & \multicolumn{1}{c}{65.49} & \multicolumn{1}{c}{40.33} & \multicolumn{1}{c|}{45.83} \\
LED\textsubscript{H} & 59.48 & 33.30 & 40.39 \\ \hline
LED\textsubscript{NH} & 45.94 & 19.68 & 28.86 \\
LED\textsubscript{H-mix} & 46.86 & 20.26 & 29.60 \\ \hline
\end{tabular}
\caption{ROUGE-1, -2 and -L content preservation results, comparing model output to the gold summaries. We evaluate our baseline models (LED\textsubscript{H} and LED\textsubscript{only-H}), along with the alternative compared configurations (LED\textsubscript{NH} and LED\textsubscript{H-mix}).}
\label{tab:evaluation-results}
\end{table}